%% file: main_cr.tex
\documentclass[conference]{IEEEtran}
\IEEEoverridecommandlockouts
\usepackage{cite}
\usepackage{amsmath,amssymb,amsfonts}
\usepackage{algorithm}
\usepackage{algorithmic}
\usepackage{pdfpages}
\usepackage{multirow}
\usepackage[export]{adjustbox} 
\usepackage{bm}
\usepackage{amsmath}
\usepackage{array}
\usepackage{booktabs}
\usepackage{footmisc}

\usepackage{amsthm}
\usepackage{amsmath}
\usepackage{amsfonts}
\usepackage{graphicx}
\usepackage{enumitem}

\theoremstyle{Definition}
\newtheorem{mydef}{Definition}
\newtheorem{mylemma}{Lemma}
\usepackage{subfigure}
\usepackage{hyperref}
\usepackage{url}
\usepackage{color}
\usepackage{xcolor}
\usepackage{colortbl}
\usepackage{xspace}
\newcommand{\model}{RAND\xspace}
\newcommand{\std}[1]{\textcolor{gray}{\scriptsize{$\pm$#1}}}
\input{math_commands.tex}

\def\BibTeX{{\rm B\kern-.05em{\sc i\kern-.025em b}\kern-.08em
    T\kern-.1667em\lower.7ex\hbox{E}\kern-.125emX}}
\begin{document}

\title{Reinforcement Neighborhood Selection for Unsupervised Graph Anomaly Detection}

\author{
\IEEEauthorblockN{Yuanchen Bei$^1$, Sheng Zhou$^1$\IEEEauthorrefmark{2}\thanks{† Corresponding author.}, Qiaoyu Tan$^2$, Hao Xu$^3$,}
\IEEEauthorblockN{Hao Chen$^4$, Zhao Li$^1$, Jiajun Bu$^1$}
\IEEEauthorblockA{$^1$ \textit{Zhejiang Provincial Key Laboratory of Service Robot, Zhejiang University, Hangzhou, China}}
\IEEEauthorblockA{$^2$ \textit{New York University Shanghai, Shanghai, China}}
\IEEEauthorblockA{$^3$ \textit{Unaffiliated, Beijing, China}}
\IEEEauthorblockA{$^4$ \textit{The Hong Kong Polytechnic University, Hong Kong SAR, China}}
\IEEEauthorblockA{yuanchenbei@zju.edu.cn, zhousheng\_zju@zju.edu.cn, qiaoyu.tan@nyu.edu, kingsleyhsu1@gmail.com,}
\IEEEauthorblockA{sundaychenhao@gmail.com, lzjoey@gmail.com, bjj@zju.edu.cn}
}


\maketitle

\begin{abstract}
\input{abstract.tex}
\end{abstract}


\section{Introduction}
\input{introduction.tex}

\begin{figure*}[tbp]
    \centering
    \includegraphics[width=\linewidth, trim=0cm 0cm 0cm 0cm,clip]{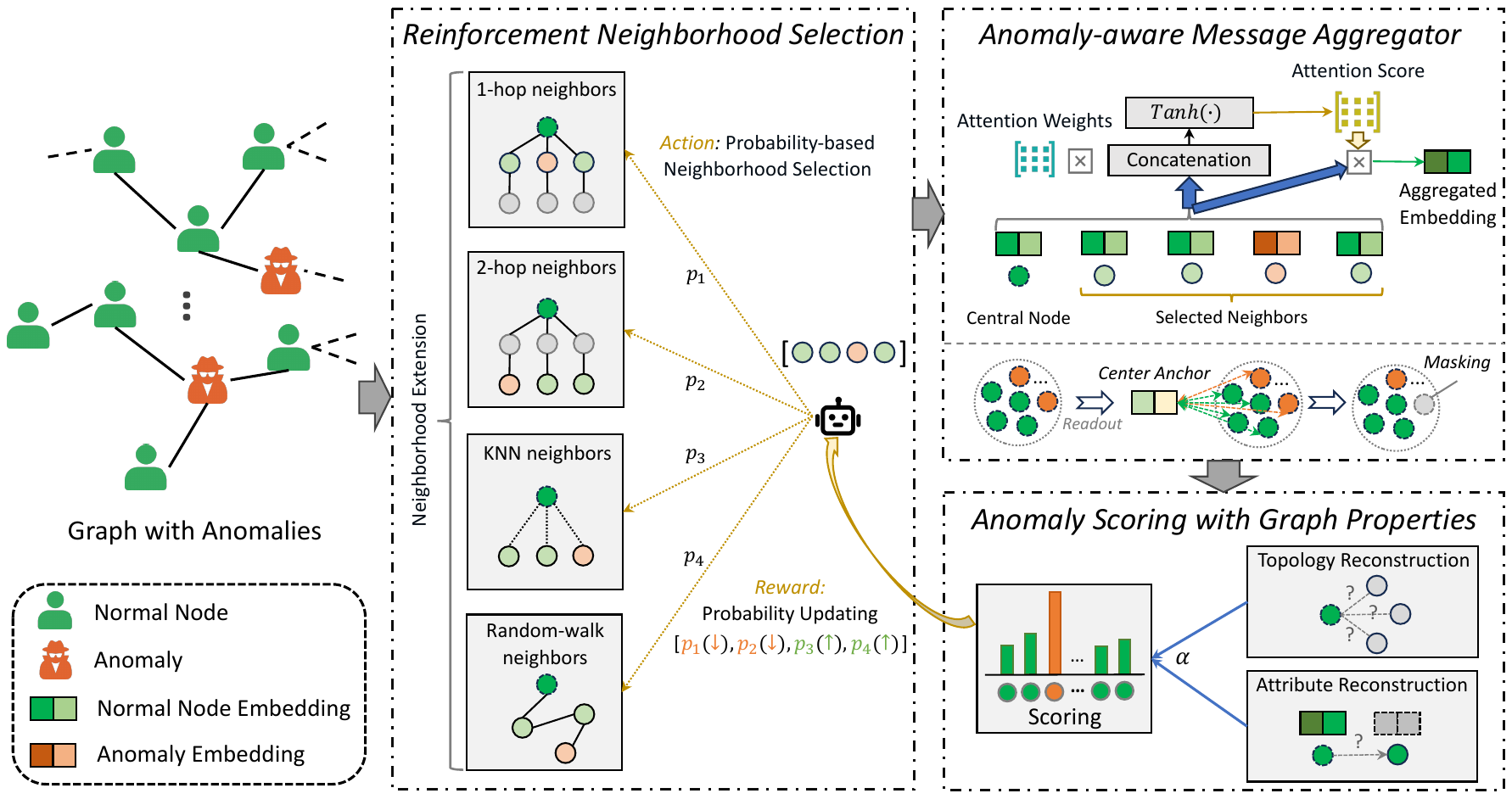}
    \caption{Network architecture of the proposed \model model. 
    \model is composed of three main parts: 
    (i) the Reinforcement Neighborhood Selection which extends the neighborhood into multiple perspectives and adaptively selects neighboring nodes by multi-arm bandit reinforcement learning.
    (ii) the Anomaly-aware Message Aggregator which modifies the traditional message aggregator to make the learned representations more distinguishable for anomalies.
    (iii) the Anomalies Scoring with Graph Properties evaluates nodes' abnormality through their ability to reconstruct the graph properties, i.e. topology and attribute. The anomaly scoring consistency between central nodes and their selected neighbors will be regarded as the reward of the neighborhood selecting action.}
    \label{fig:framework}
    \vspace{-0.7em}
\end{figure*}

\section{Related Works}
\input{related.tex}

\section{Problem Statement}
\input{preliminaries.tex}

\section{Methodology}
\input{method.tex}

\section{Experiments}
\input{experiment.tex}

\section{Conclusion}
\input{conclusion.tex}

\section{ACKNOWLEDGEMENT}
This work is supported in part by the National Natural Science Foundation of China (Grant No. 62106221, 61972349), Zhejiang Provincial Natural Science Foundation of China (Grant No. LTGG23F030005), and Ningbo Natural Science Foundation (Grant No. 2022J183).

\bibliographystyle{IEEEtran}
\bibliography{IEEEabrv,ref}

\end{document}

%% file: math_commands.tex
\usepackage{amsmath,amsfonts,bm}

\def\eqref#1{equation~\ref{#1}}

\def\1{\bm{1}}

\DeclareMathAlphabet{\mathsfit}{\encodingdefault}{\sfdefault}{m}{sl}
\SetMathAlphabet{\mathsfit}{bold}{\encodingdefault}{\sfdefault}{bx}{n}

\def\gC{{\mathcal{C}}}

\def\gE{{\mathcal{E}}}
\def\gF{{\mathcal{F}}}
\def\gG{{\mathcal{G}}}

\def\gL{{\mathcal{L}}}

\def\gN{{\mathcal{N}}}
\def\gO{{\mathcal{O}}}
\def\gP{{\mathcal{P}}}

\def\gV{{\mathcal{V}}}
\def\gW{{\mathcal{W}}}



%% file: abstract.tex
Unsupervised graph anomaly detection is crucial for various practical applications as it aims to identify anomalies in a graph that exhibit rare patterns deviating significantly from the majority of nodes. 
Recent advancements have utilized Graph Neural Networks (GNNs) to learn high-quality node representations for anomaly detection by aggregating information from neighborhoods. 
However, the presence of anomalies may render the observed neighborhood unreliable and result in misleading information aggregation for node representation learning.
Selecting the proper neighborhood is critical for graph anomaly detection but also challenging due to the absence of anomaly-oriented guidance and the interdependence with representation learning. 
To address these issues, we utilize the advantages of reinforcement learning in adaptively learning in complex environments and propose a novel method that incorporates \underline{\textit{R}}einforcement neighborhood selection for unsupervised graph \underline{\textit{AN}}omaly \underline{\textit{D}}etection (RAND).
\model begins by enriching the candidate neighbor pool of the given central node with multiple types of indirect neighbors.
Next, \model designs a tailored reinforcement anomaly evaluation module to assess the reliability and reward of considering the given neighbor. 
Finally, \model selects the most reliable subset of neighbors based on these rewards and introduces an anomaly-aware aggregator to amplify messages from reliable neighbors while diminishing messages from unreliable ones.
Extensive experiments on both three synthetic and two real-world datasets demonstrate that \model outperforms the state-of-the-art methods.

%% file: introduction.tex
Unsupervised graph anomaly detection aims to discover the graph anomalies that exhibit rare patterns from the majority in an unsupervised manner.
It has gained increasing attention from both academia and industry in recent years for the sparsity of anomaly labels and the significant real-world applications~\cite{ding2019deep,dagad}, such as financial fraud detection~\cite{wang2019semi}, social rumor detection~\cite{sun2022rumor}, and computer network intrusion detection~\cite{zhou2021hierarchical}.
Benefiting from the success of Graph Neural Networks (GNNs) in learning effective node representations~\cite{kipf2016semi,hamilton2017inductive}, recent advances have widely adopted GNNs to learn high-quality node embeddings for unsupervised graph anomaly detection~\cite{ma2021comprehensive}.

\begin{figure}[tbp]
    \centering
    \includegraphics[width=\linewidth, trim=0cm 0cm 0cm 0cm,clip]{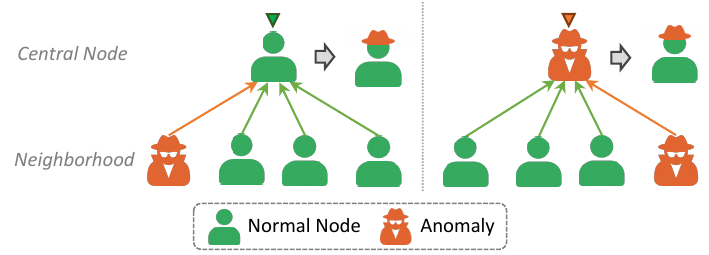}
    \caption{A toy example of the negative impact on central normal nodes (left) and anomaly nodes (right) by unreliable neighborhood in GNN-based representation learning and anomaly detection.}
    \vspace{-1em}
    \label{fig:toy-example}
\end{figure}

The existing GNN-based methods for graph anomaly detection have mostly learned representation by aggregating information from neighborhoods.
This is motivated by the \textit{homophily assumption} that nodes close in the graph tend to exhibit similar patterns as the central nodes~\cite{mcpherson2001birds}, which is widely observed in real-world graphs.
However, in the graph anomaly detection scenario, the presence of anomalies can disrupt this homophily assumption and render the \textbf{neighborhood unreliable}.
More specifically, the misleading information aggregated from the unreliable neighborhood will result in suboptimal representation learning and anomaly detection performance.
For the normal nodes, as illustrated in the left part of Figure \ref{fig:toy-example}, they suffer from aggregating noisy information from the anomaly nodes included in their neighborhood, resulting in inaccurate node modeling.
For the anomaly nodes, as illustrated in the right part of Figure \ref{fig:toy-example}, they can camouflage themselves as normal ones by establishing connections with numerous regular nodes~\cite{dou2020enhancing} and aggregating conventional messages from their normal neighborhood. 
Consequently, the observed neighborhood becomes less reliable due to the presence of anomalies, making it crucial to carefully select appropriate neighborhoods for accurate representation learning and effective anomaly detection.

Although important, selecting appropriate neighborhood in unsupervised graph anomaly detection poses the following significant challenges:
First, the unsupervised setting prevents us from accessing the nodes' ground-truth labels, which makes it lacks \textbf{anomaly-oriented guidance} for neighborhood selection~\cite{ding2019deep,liu2022bond}.
Second, the appropriate selection of neighborhoods and the high-quality node representation learning are \textbf{mutually dependent}. However, in the early stage of model training, both the reliability of neighborhoods and the effectiveness of representation learning are inadequate.
Therefore, it is necessary to dynamically adjust the neighborhood selection strategy during the training process to enhance the quality of representation learning for anomaly distinguishing.
Lastly, anomalies in the real world exhibit a wide range of diversity, with significant variations in abnormal patterns across different scenarios~\cite{liu2022bond}. For instance, anomalies in social networks differ greatly from those in power networks. Consequently, the neighborhood selection process should \textbf{autonomously adapt} to accommodate different graphs.

To tackle the above challenges, in the paper, we propose a novel method that incorporates \textbf{R}einforcement neighborhood selection for unsupervised graph \textbf{AN}omaly \textbf{D}etection (\textbf{RAND}).
Reinforcement learning (RL) has demonstrated superior capability in adaptive learning in complex environments, which conveniently fulfills the demands of neighbor selection mentioned above.
Specifically, \model first extends the observed neighborhood to several noteworthy groups that may potentially benefit the anomaly detection, including 1-hop, 2-hop, high-order, and attribute-based neighborhoods.
Then, \model follows a reinforcement learning paradigm and treats the dynamic probability-based neighborhood selection as the \textit{action}.
The quality of selection is evaluated by the consistency of anomaly scores between central nodes and selected neighborhoods, which serves as the \textit{reward} of the action.
Subsequently, we further design an anomaly-aware message aggregator to fully leverage the information contained in the selected neighborhoods.
Finally, \model adopts reconstruction on the graph properties for model training and anomaly scoring in an unsupervised way. 
Extensive experiments conducted on both synthetic and real-world datasets demonstrate that \model outperforms state-of-the-art models.
The main contributions of this paper are organized as follows:
\begin{itemize}[leftmargin=*]
    \item We highlight the negative impact of anomaly nodes on neighborhood reliability, which is crucial for existing GNN-based graph anomaly detection methods.
    \item We propose \model, a novel unsupervised graph anomaly detection method with reinforcement neighborhood selection and anomaly-aware message aggregation.
    \item We conduct extensive experiments on both widely-used synthetic and real-world datasets, which show that \model outperforms the state-of-the-art methods.
\end{itemize}

%% file: related.tex
\subsection{Unsupervised Graph Anomaly Detection}
Traditional methods for unsupervised graph anomaly detection mainly focus on feature engineering or directly utilize instance attributes with shallow neural networks, e.g., SCAN~\cite{xu2007scan} and MLPAE~\cite{sakurada2014anomaly}, regardless of the instance relationship modeling. 
Due to the success of GNNs, recent works propose to consider both the attribute and topology abnormal patterns for anomaly mining and achieve state-of-the-art unsupervised graph anomaly detection performance.
Among them, GAAN~\cite{chen2020generative}, ALARM~\cite{peng2020deep}, and AAGNN~\cite{zhou2021subtractive} try to improve the basic GNNs with enhanced unsupervised modeling training framework or representation learning procedure to strengthen the model sensitivity to anomalies.
Another type of model adopts the deep graph autoencoder, e.g., GCNAE~\cite{kipf2016variational}, Dominant~\cite{ding2019deep}, AnomalyDAE~\cite{fan2020anomalydae}, and ComGA~\cite{luo2022comga}, based on the imbalanced number of nodes, the model learns the patterns of most normal nodes in the graph, thus the anomalies cannot be well reconstructed. The reconstruction error is used to evaluate whether a node is an anomaly.
Recently, with the wide application of self-supervised learning in graphs, the methods based on graph contrastive learning become another category of models, i.e. CoLA~\cite{liu2021anomaly}, ANEMONE~\cite{jin2021anemone}, SL-GAD~\cite{zheng2021generative}, and Sub-CR~\cite{zhang2022subcr}, in which the magnitude of difference between positive and negative sample pairs designed for node identification is utilized to evaluate a node's abnormality.

Most of the methods adopt vanilla GNN message passing to unsupervised learn node representations for anomaly distinguishing. 
Nevertheless, the neighborhood unreliable caused by anomaly nodes will bring noise information during the message passing and is yet to be addressed well.

\subsection{Heterophily-based GNNs}
For the powerful graph modeling ability, many GNNs have been proposed in recent years, such as GCN~\cite{kipf2016semi}, GraphSage~\cite{hamilton2017inductive}, and GAT~\cite{veličković2018graph}. They implicitly assume that connected nodes have similar behaviors (i.e. attributes or labels), which is typically called the \textit{homophily assumption}~\cite{mcpherson2001birds}. 

Recently, another type of graph modeling method called \textit{heterophily-based GNN} is proposed for representation learning under heterophilic connections. 
These methods can be mainly divided into two kinds:
(1) The first kind of approach is to mix multiple types of information to minimize the passing and proportion of heterophilic information, such as MixHop~\cite{abu2019mixhop} and H2GCN~\cite{zhu2020beyond}.
(2) The second approach is to model homophilic information and heterophilic information separately by label information, and then using a fusion module to obtain the final node representation to alleviate information conflicts, such as LINKX~\cite{lim2021large} and GloGNN~\cite{li2022finding}.
Recently,~\cite{shi2022h2} identifies the heterophilic connections with the supervision of labeled nodes for fraud detection.
However, in unsupervised graph anomaly detection, it lacks effective label signals and particular anomaly modeling for them to work.

\subsection{Deep Reinforcement Learning on Graphs}
With the success of deep reinforcement learning (RL) in various research fields, such as robotics~\cite{kober2013reinforcement} and games~\cite{lanctot2017unified}.
Recently, some works have started to explore the application of RL for graph data mining~\cite{nie2023reinforcement}.

The fundamental framework involves introducing deep RL to guide the learning process of GNNs, obtaining better node representations.
Among them, Policy-GNN~\cite{lai2020policy} chooses the suitable number of GNN aggregation layers for different nodes with RL.
Then, ANS-GT~\cite{ansgt} introduces multi-arm bandits based on the attention matrix for informative node sampling in graph transformer, which inspired our model design.
CARE-GNN~\cite{dou2020enhancing} proposes RL-improved GNN based on label-aware similarity measurement for dissimilar neighbor filtering in supervised fraud detection.
However, it is difficult for them to effectively confront the challenges associated
with unsupervised graph anomaly detection.

%% file: preliminaries.tex
\noindent \textbf{Notations.} Let $\gG = (\bm{A}, \bm{X})$ be an attributed graph with a node set $\gV = \{v_{1}, v_{2}, ..., v_{n}\}$ and the edge set $\gE$, where $|\gV| = n$. $\bm{A} \in \mathbb{R}^{n \times n}$ denotes the graph adjacency matrix, $A_{i,j} = 1$ indicates that there is an edge between node $v_{i}$ and node $v_{j}$, and otherwise $A_{i,j} = 0$. $\bm{X} \in \mathbb{R}^{n \times d}$ denotes the node attribute matrix, the $i$-th row $\bm{x}_{i} = \bm{X}[i,:] \in \mathbb{R}^{d}$ indicates the attribute vector of $v_{i}$ with $d$ dimensional representation.
$\gN_{i}$ is the neighbor set of a central node $i$ in the graph $\gG$.

\begin{mydef}
    \textbf{Unsupervised Graph Anomaly Detection}: Given an abnormal attributed graph $\gG = (\bm{A}, \bm{X})$ containing $n$ node instances, and $b$ of them are anomalies ($b \ll n$), whose attributes, connections or behaviors are different from most other normal nodes. 
    The target of graph anomaly detection is to learn a model $\gF(\cdot): \mathbb{R}^{n \times n} \times \mathbb{R}^{n \times d} \rightarrow \mathbb{R}^{n}$ in the unsupervised manner that outputs anomaly score vector $\bm{S}$ to measure the degree of abnormality of nodes, where a larger score means a higher abnormality. 
\end{mydef}

%% file: method.tex
\subsection{Overall Framework of \model}
To select appropriate neighborhoods and fully leverage the information within the selected neighborhoods to learn anomaly-distinguishing representations, \model consists of three main modules: 
Reinforcement Neighborhood Selection, Anomaly-aware Message Aggregator, and Anomaly Scoring with Graph Properties.
First, \textit{Reinforcement Neighborhood Selection} extends the concept of the directly connected neighborhood from various perspectives and adaptively selects suitable neighboring nodes with dynamic selection probabilities by multi-arm bandit reinforcement learning.
Then, \textit{Anomaly-aware Message Aggregator} modifies the traditional message aggregator with more distinguishing operators to make the aggregation better utilize the information of the selected neighboring nodes.
Finally, \textit{Anomaly Scoring with Graph Properties} evaluates the degree of abnormality of each node with the graph fundamental properties (i.e. topology and attribute).
According to the scoring consistency between central nodes and their selected neighborhoods, \model provides reward feedback to dynamically update the selection probabilities.
The overall framework of \model is illustrated in Figure \ref{fig:framework} and the details of each part are introduced as follows.

\subsection{Reinforcement Neighborhood Selection}
Due to the messages passed into the center node are of significance for the quality of the learned representation~\cite{dai2022towards}, the insight of reinforcement neighborhood selection is to mine 
appropriate neighboring nodes for central nodes rather than limit to the directly connected neighbors with potential noisy and polluted information.
On account of the unsupervised setting, we cannot directly select neighboring nodes using label information.
Inspired by \cite{ansgt} for neighborhood sampling, we found that this scenario satisfies the adversarial conditions to apply the multi-armed bandit RL for adaptively selecting suitable neighborhoods for each central node.

\textbf{Applicability analysis}:
In multi-armed bandit RL, it highlights two conditions~\cite{ansgt,auer2002nonstochastic}:
1) The impact of the action can \textit{be varied over time}.
2) The rewards for the action are \textit{not independent random variables} throughout training.
Our scenario is consistent with its assumptions for: 
1) It is intuitive that the influence of selected neighboring nodes on the anomaly detection performance can \textit{shift over time}.
2) The rewards based on the anomaly scores are linked to the model training process.
These two properties satisfy the adversarial premise of utilizing multi-armed bandit RL.


\textbf{\textit{Action}}: Considering different selection strategies can obtain neighborhoods with different preferences, we utilize multi-armed bandit to determine the preferences for the selection strategies.
Let $\bm{\gW}^{t} = (w^{t}_{1}, ..., w^{t}_{K})$ be the adaptive weight vector in training iteration $t$, where $w^{t}_{k}$ is the weight corresponding to the $k$-th selection strategy in iteration $t$, and $K$ is the number of selection heuristics.
Then, the weight vector $\bm{\gW}^{t}$ will be mapped to a strategy-grained probability vector $\bm{\gP}^{t} = (p^{t}_{1}, ..., p^{t}_{K})$, where $p^{t}_{k}\in [p_{min}, 1]$, and $p_{min} \geq 0$ is a controllable constant that constrains the lower bound of probability, which is set to a default value of 0.05.
\model adaptively selects candidate nodes for aggregating based on the probability vector as the reinforcement action (bandit) in the environment of an abnormal graph.
For each center node, we consider the fine-grained selection probability matrix $\bm{Q}^{t}\in\mathbb{R}^{K \times n}$. Specifically, $Q^{t}_{k,j}$ denotes the $k$-th selection strategy’s preference on selecting node $j$ in iteration $t$ and $\bm{Q}^{t}_{k}$ is normalized where $\sum_{j=1}^{n}Q^{t}_{k,j}=1$.
Note that \model is a general framework and is not restricted to a certain set of selection heuristics. Here we adopt four representative selection heuristics~\cite{ansgt}:
\begin{itemize}[leftmargin=*]
    \item \textbf{1-hop neighbors}: The directly connected neighbors sampled from the normalized adjacency matrix.
    \item \textbf{2-hop neighbors}: The indirectly connected nodes sampled from the power of normalized adjacency matrix.
    \item \textbf{KNN neighbors}: The nodes whose relationship is based on the cosine similarity of node attributes.
    \item \textbf{Random-walk neighbors}: The nodes with informative high-order patterns. The Personalized PageRank~\cite{wang2020personalized} is utilized for generating the random-walk weight.
\end{itemize}

Given strategy-grained probability $p^{t}$ and fine-grained probability $\bm{Q}^{t}$, the final selecting probability for node $i$ at training epoch $t$ is:
\begin{equation}
    \Phi^{t}_{i} = \sum_{k=1}^{K}p_{k}^{t} \cdot Q_{k,i}^{t},
\end{equation}

\textbf{\textit{Reward}}: Nevertheless, \textit{the follow-up question is how to update the selecting probability}. For a central node, a good selection should bring more consistent and homophilic information to it.
According to this philosophy, we design the reward of the selection action based on the anomaly score consistency $\bm{\gC}^{t}$ between the central node and its selected neighboring nodes, of which the anomaly scoring details with be introduced in the following subsection D.

Given the anomaly score $y_{i}$ for the center node $i$, we first obtain the score vector $\bm{Y}_{\gN_{i}}$ of its last selected neighboring nodes and calculate the score similarity distribution between $i$ and $\gN_{i}$ as follows:
\begin{equation}
    \bm{C}^{t}_{i} = Softmax(\frac{1}{abs(y_{i}-y_{j})+\epsilon}), j \in \gN_{i}.
\end{equation}
Then, we give the reward based on the distribution similarity between $\bm{C}^{t}_{i}$ and $\Phi^{t}_{i}$, in which a higher similarity indicates the selecting way is more meaningful and thus higher reward will be given.
Formally, the reward is calculated by a dot product scheme as follows:
\begin{equation}
    r^{t}_{k} = \frac{1}{n}\sum_{i=1}^{n} \bm{C}_{i}^{t}\cdot \frac{p^{t}_{k}\cdot \bm{Q}_{k,\gN_{i}}^{t}}{\bm{\Phi}_{\gN_{i}}^{t}},
\end{equation}
where $r^{t}_{k}$ is the reward vector of different selection strategies.
Then the selection strategy weight can be updated as follows:
\begin{equation}
    w^{t+1}_{k} = w^{t}_{k} e^{(\frac{p_{min}}{2})(r_{k}+\frac{1}{p^{t}_{k}})\delta_{1}\sqrt{\frac{ln(n/\delta_{2})}{KT}}},
    \label{eq:weight_update}
\end{equation}
where $T$ is the update internal, $\delta_{1}, \delta_{2}$ are controllable parameters.
Note that we calculate the reward $U$ after warm-up epochs $U$ (we set $U=3$ in our experiments) to ensure the model has the basic ability to distinguish anomalies.

\begin{figure}[tbp]
    \centering
    \includegraphics[width=\linewidth, trim=0cm 0cm 0cm 0cm,clip]{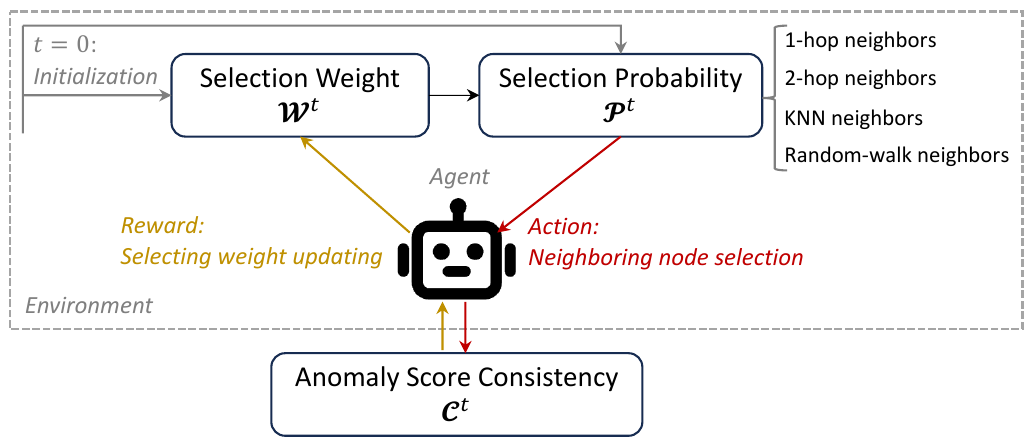}
    \caption{The overall pipeline of the reinforcement neighborhood selection.}
    \label{fig:rl-pipeline}
\end{figure}

Finally, the strategy-grained selection probability matrix for $k$-th strategy can be updated as follows:
\begin{equation}
    p_{k}^{t+1} = (1-Kp_{min})\cdot\frac{w_{k}^{t+1}}{SW^{t+1}}+p_{min},
    \label{eq:p_update1}
\end{equation}
\begin{equation}
    SW^{t+1} = \sum_{k=1}^{K} w^{t+1}_{k}.
    \label{eq:p_update2}
\end{equation}
We iterate the \model training with the updated strategy-grained selection probability matrix $p_{k}^{t+1}$. 
Figure \ref{fig:rl-pipeline} illustrates the general process of reinforcement neighborhood selection.

\subsection{Anomaly-aware Message Aggregator}
After we obtain the selected neighborhood, our target is to learn the anomaly-distinguishing representation with the homophilic message under unsupervised settings. The insight of the designed Anomaly-aware Message Aggregator $f_{ama}(\cdot)$ is to: (i) Mask potential anomaly nodes, disallowing message passing for this subset of nodes to expand their representation dissimilarity with the majority of normal nodes; (ii) Modify the message aggregation function in previous works by using operations that provide more message differentiation capability to aggregate more homophilic patterns for the central node.
With the above insights, $f_{ama}(\cdot)$ contains the two main stages: \textit{center node masking} and \textit{distinguishing message passing}.

\textbf{Center Node Masking.} We first project nodes' sparse features $\bm{X}$ into the latent space embeddings $\bm{E}$ with a two-layer MLP. Then, we mask potential anomaly nodes to prevent them from aggregating messages from a large number of normal nodes around them, as this could even help them to disguise themselves. We define a \textit{center anchor} of the overall graph, those nodes that are farther from the center anchor are considered as the potential anomalies, which will be banned from message passing.
Specifically, the representation of the center anchor $\bm{E}_{ca}$ is obtained by a readout function:
\begin{equation}
    \bm{E}_{ca}=Readout(\bm{E}),
\end{equation}
where $Readout(\cdot)$ can be a kind of pooling operation (such as min, max, mean, and weighted pooling~\cite{liu2021anomaly}), here we use the mean pooling for simplicity.
Then, with a given mask rate $mr \geq 0$, we mask those nodes that are farthest from the center anchor $\bm{E}_{ca}$, of which the mask nodes can be calculated as:
\begin{equation}
    M_{id} = Top_{(mr\cdot|\gV|)}[argsort_{max}(||\bm{E}_{i}-\bm{E}_{ca}||_{2}^{2}, i \in \gV)],
    \label{eq:mask_ids}
\end{equation}
where $||\cdot||_{2}^{2}$ is the Euclidean distance and $M_{id}$ is the set of masked center nodes that will be skipped from the following message passing.
For these masked nodes, their encodings will be identified with their original representation:
\begin{equation}
    \bm{H}_{i} = \bm{E}_{i}, i\in M_{id},
\end{equation}
where $\bm{H}_{i}$ is the graph encoding of center node $i$.

\textbf{Distinguishing Message Passing.}
The traditional message passing functions are insufficient to distinguish neighborhood messages~\cite{hamilton2017inductive,veličković2018graph} since they consider all information from the neighborhood as homophilic messages and assimilate them.
To address this, we modify the message passing function with fine-grained and more distinguishing operators. The distinguishing message passing contains the following main operations: 
(i) We calculate the attention score for the central node and its entire selected neighborhood to evaluate one neighbor's consistency not only with the central node but also with the other neighbors of the central node, rather than compute between the central node and each of its neighbors separately in previous research.
(2) We extend the attention vector to an attention matrix to obtain the fine-grained dimensional attention score and utilize $Tanh(\cdot)$ as the activation function rather than $Softmax(\cdot)$ to further distinguishing the negative messages for the center node.
Formally, the process of message passing can be described as follows:
\begin{equation}
    \bm{H}_{i} = (\bm{att}_{i,i} \odot \bm{E}_{i} + \sum_{j \in \gN_{i}} \bm{att}_{i,j} \odot \bm{E}_{j})\bm{W}_{mp},
\end{equation}
\begin{equation}
    \bm{att}_{i} = Tanh(Concat[\bm{E}_{i}, \{\bm{E}_{j}, j \in \gN_{i}\}]\bm{W}_{att}),
\end{equation}
where $\bm{W}_{mp}$ and $\bm{W}_{att}$ are trainable parameters, $\bm{att}_{i} \in \mathbb{R}^{|\gN|\times d}$ is the \textit{attention score matrix} of the center node $i$, and $\bm{att}_{i,j}\in \mathbb{R}^{d}$ is the score vector of node $j$ to node $i$. 

\subsection{Anomaly Scoring with Graph Properties}
The anomalies exhibit inconsistent patterns that deviate from the majority of the graph~\cite{akoglu2015graph,ding2019deep}.
Therefore, with the well-trained node representations, we conduct anomaly scoring to mining anomalies based on the reconstruction ability of nodes on the main graph properties, i.e. topology and attribute.

\textbf{Topology Reconstruction}.
The topology is an important property of nodes in a graph, thus the reconstruction capability is a factor for potential anomalies mining.
For the learned representation $\bm{H}$, the topology reconstruction function $g_{t}(\cdot)$ reconstructs the graph adjacency matrix as follows:
\begin{gather}
    \hat{\bm{A}}=g_{t}(\bm{H})=\bm{H}\cdot\bm{H}^{T},\\
    \bm{D}_{topo} = ||\bm{A}-\hat{\bm{A}}||_{2}^{2},
\end{gather}
where $g_{t}(\cdot)$ is a simple inner product operation between $\bm{H}$ and its self-transposition $\bm{H}^{T}$ to ensure the reconstruction efficiency, and $\bm{D}_{topo} \in \mathbb{R}^{n}$ is the topology reconstruction error vector of the nodes, of which each element $D_{topo, i}$ is viewed as a factor for topological anomaly scoring of node $i$.

\textbf{Attribute Reconstruction}.
The attribute is also the key property of nodes that reflects the characteristics of the node itself. The attribute reconstruction capability is another evaluation factor to mine anomalies due to the inconsistent attribute is difficult to be reconstructed.
Given the representation $\bm{H}$, an attribute reconstruction function $g_a(\cdot)$ is implemented to reconstruct the nodes' original attribute as follows:
\begin{gather}
    \hat{\bm{X}}=g_{a}(\bm{H}, \bm{A}),\\
    \bm{D}_{attr} = ||\bm{X}-\hat{\bm{X}}||_{2}^{2},
\end{gather}
where $g_{a}(\cdot)$ can be a kind of graph encoder, such as MLP and GCN~\cite{kipf2016semi}, and $\bm{D}_{attr} \in \mathbb{R}^{n}$ is the attribute reconstruction error vector of the nodes, of which each element $D_{attr, i}$ can be utilized for attributed anomaly scoring of node $i$.

Therefore, we utilize the above two aspects of graph property reconstruction errors for the final anomaly scoring.
Here the scoring process of node $i$ is calculated as:
\begin{equation}
\begin{aligned}
    score(i) &= (1-\alpha) \cdot D_{topo, i} + \alpha \cdot D_{attr, i}\\ &= (1-\alpha) \cdot||\bm{a}_{i}-\hat{\bm{a}}_{i}||_{2}^{2}+ \alpha \cdot||\bm{x}_{i}-\hat{\bm{x}}_{i}||_{2}^{2},
\end{aligned}
\label{eq:ano_score}
\end{equation}
where $\alpha$ is a balanced parameter between the topology reconstruction and attribute reconstruction. Nodes with larger scores are more likely to be considered as anomalies.

\subsection{Objective Function}
To jointly learn the topology and attribute reconstruction errors for model training, the objective function of \model can be formulated as:
\begin{equation}
{\gL}_{topo} = \frac{1}{|\gV|}\sum ||\bm{A}-\hat{\bm{A}}||_{2}^{2},
\end{equation}
\begin{equation}
{\gL}_{attr} = \frac{1}{|\gV|}\sum ||\bm{X}-\hat{\bm{X}}||_{2}^{2},
\end{equation}
\begin{equation}
    \gL = (1-\alpha) \cdot{\gL}_{topo} + \alpha\cdot{\gL}_{attr} + \lambda \cdot ||\theta||^{2}_{2},
\end{equation}
where $\theta$ is the parameter set of \model, $\alpha$ is the balanced hyper-parameter the same as mentioned Eq.(\ref{eq:ano_score}), and $\lambda$ is the regularizer parameter.

\subsection{Model Analysis}

\textbf{Convergence Analysis.}
Let $r^{*}$ be the reward value of the optimal solution, $r_{k}$ be the reward value of the $k$-th selection strategy, $p_{k}^{*}$ be the selection probability that achieves the optimal value $r^{*}$ for $r_{k}$.

\begin{mylemma}
    For any given selection strategy $k$, the expected value of $w_{k}^{t}$ is $w_{k}^{*}$.
\end{mylemma}
\textit{Prove}: Consider Eq.(\ref{eq:weight_update}), taking the expected value, we have:
\begin{equation}
\begin{aligned}
\mathbb{E}[w_{k}^{t+1}] &= \mathbb{E}[w_{k}^{t} e^{(\frac{p_{min}}{2})(r_{k}+\frac{1}{p_{k}^{t}})\delta_{1}\sqrt{\frac{ln(n/\delta_{2})}{KT}}}] \\
&= \mathbb{E}[w_{k}^{t}] \mathbb{E}[e^{(\frac{p_{min}}{2})(r_{k}+\frac{1}{p_{k}^{t}})\delta_{1}\sqrt{\frac{ln(n/\delta_{2})}{KT}}}].
\end{aligned}
\label{eq:expected}
\end{equation}
According to the Jensen's inequality for exponential functions and its convexity property, we have:
\begin{equation}
\begin{aligned}
\mathbb{E}[e^{(\frac{p_{min}}{2})(r_{k}+\frac{1}{p_{k}^{t}})\delta_{1}\sqrt{\frac{ln(n/\delta_{2})}{KT}}}] &\geq e^{\mathbb{E}[(\frac{p_{min}}{2})(r_{k}+\frac{1}{p_{k}^{t}})\delta_{1}\sqrt{\frac{ln(n/\delta_{2})}{KT}}]}\\
&\geq e^{(\frac{p_{min}}{2})(\mathbb{E}[r_{k}]+\frac{1}{\mathbb{E}[p_{k}^{t}]})\delta_{1}\sqrt{\frac{ln(n/\delta_{2})}{KT}}}.
\end{aligned}
\label{eq:jensen}
\end{equation}
Substituting Eq.(\ref{eq:jensen}) into Eq.(\ref{eq:expected}) along with the definition of $p^{*}_{k}$, we get:
\begin{equation}
\begin{aligned}
\mathbb{E}[w_{k}^{t+1}] &\geq \mathbb{E}[w_{k}^{t}] e^{(\frac{p_{min}}{2})(\mathbb{E}[r_{k}]+\frac{1}{\mathbb{E}[p_{k}^{t}]})\delta_{1}\sqrt{\frac{ln(n/\delta_{2})}{KT}}} \\
&= \mathbb{E}[w_{k}^{t}] \frac{w_{k}^{*}}{\mathbb{E}[w_{k}^{t}]}.
\end{aligned}
\end{equation}
Therefore, we have $\mathbb{E}[w_{k}^{t+1}] \geq w_{k}^{*}$ that the expected value of $w_{k}^{t}$ is $w_{k}^{*}$, which proves Lemma 1.
\begin{mylemma}
    For any given selection strategy $k$, the expected value of $p_{k}^{t}$ approaches $p_{k}^{*}$.
\end{mylemma}
\textit{Prove}: Taking the expected value of the updating of $p_{k}^{t}$ in Eq.(\ref{eq:p_update1}) and Eq.(\ref{eq:p_update2}), we have:
\begin{equation}
\begin{aligned}
\mathbb{E}[p_{k}^{t+1}] &= (1-Kp_{min})\cdot\mathbb{E}[\frac{w_{k}^{t+1}}{SW^{t+1}}]+p_{min} \\
&= (1-Kp_{min})\cdot\frac{\mathbb{E}[w_{k}^{t+1}]}{\mathbb{E}[SW^{t+1}]}+p_{min},
\end{aligned}
\end{equation}
Then, since the expected value of $w_{k}^{t}$ approaches $w_{k}^{*}$, we have:
\begin{equation}
    \lim_{t\rightarrow \infty} \mathbb{E}[w_{k}^{t}] = w_{k}^{*},
\end{equation}
\begin{equation}
    \lim_{t\rightarrow \infty} \mathbb{E}[SW^{t}] = K w_{k}^{*},
\end{equation}
\begin{equation}
\begin{aligned}
\lim_{t\rightarrow \infty} \mathbb{E}[p_{k}^{t+1}] &= (1-Kp_{min})\cdot\frac{\lim_{t\rightarrow \infty} \mathbb{E}[w_{k}^{t+1}]}{\lim_{t\rightarrow \infty} \mathbb{E}[SW^{t+1}]}+p_{min} \\
&= (1-Kp_{min})\cdot\frac{w_{k}^{*}}{K w_{k}^{*}}+p_{min} \\
&= p_{k}^{*}.
\end{aligned}
\end{equation}
Here, Lemma 2 is proved.

\textbf{Convergence Guarantee}.
From the proven lemmas, we have:
Firstly, considering the selection probability $p_{k}^{t}$ of each selection strategy $k$, it can approaches $p_{k}^{*}$. Therefore, in finite steps, each selection strategy will be sufficiently learned.
Secondly, since the weight $w_{k}^{t}$ of each selection strategy approaches the optimal value $w_{k}^{*}$, each selection strategy will be selected enough times in finite steps to obtain a reward value $r_{k}$ that is close enough to the optimal value $r^{*}$. Thus, \model can converge to the optimal solution within finite steps.

\textbf{Complexity Analysis.}
The main complexity of \model is from three parts: neighborhood selection, message aggregator, and anomaly scoring.
Specifically, 1) the time complexity of the neighborhood selection is $\gO(K\cdot n)$, where $K \ll n$.
2) the complexity of $f_{ama}(\cdot)$ is $\gO(M\cdot n)$, where $M \ll n$ represents the sampled size of the central node which is set to 20 as default.
3) the anomaly scoring decoders $g_{t}(\cdot)$ and $g_{a}(\cdot)$ has the complexity of $\gO(n^2)$ and $\gO(n\cdot d)$, respectively.
Therefore, the overall complexity of \model is $\gO\left((K+M)\cdot n + n^2\right)$.

%% file: experiment.tex

\subsection{Experimental Settings}
\subsubsection{\textbf{Datasets}} 
We adopt five widely-used datasets to verify the effectiveness of \model, including three synthetic datasets: Cora, Citeseer, and Flickr, and two real-world datasets: Weibo, and Reddit.
The dataset details are introduced as follows and the statistics of the datasets are shown in Table \ref{tab:stats}.

\textbf{Synthetic datasets}:
(1) \textbf{Cora} \footnote{\url{https://linqs.soe.ucsc.edu/datac}\label{data_addr1}}~\cite{sen2008collective} is a classical citation network consisting of 2,708 scientific publications (contains 150 injected anomalies) along with 5,429 links between them.
(2) \textbf{Citeseer}\textsuperscript{\ref{data_addr1}}~\cite{sen2008collective} is also a citation network consisting of 3,327 scientific publications (contains 150 injected anomalies) with 4,732 links.
(3) \textbf{Flickr}\footnote{\url{http://socialcomputing.asu.edu/pages/datasets}}~\cite{tang2009relational} is a social network dataset acquired from the image hosting and sharing website Flickr. In this dataset, 7,575 nodes denote the users (contains 450 injected anomalies), and 239,738 edges represent the following relationships between users.

\textbf{Real-world datasets}:
(1) \textbf{Weibo}\footnote{\url{https://github.com/zhao-tong/Graph-Anomaly-Loss/tree/master/data/weibo_s}}~\cite{zhao2020error} is a user-posts-hashtag graph from Tencent-Weibo platform, which collects information from 8,405 users (contains 868 suspicious users). The provided user-user graph is used, which connects users who used the same hashtag.
(2) \textbf{Reddit}\footnote{\url{http://files.pushshift.io/reddit}}~\cite{kumar2019predicting} is a user-subreddit graph from a social media platform, Reddit, which consists of one month of user posts on subreddits. The 1,000 most active subreddits and the 10,000 most active users (containing 366 banned users) are extracted. We convert it to a user-user graph for experiments, which connects users who have edited the same subreddits.

\begin{table}[htbp]
  \centering
  \small
  \caption{Statistics of the experimental datasets.}
  \resizebox{0.85\linewidth}{!}{
    \begin{tabular}{l|c|c|c|c}
    \toprule
    Dataset & \# nodes & \# edges & \# attributes & \# anomalies \\
    \midrule
    Cora  & 2,708 & 5,429 & 1,433 & 150 \\
    Citeseer & 3,327 & 4,732 & 3,703 & 150 \\
    Flickr & 7,575 & 239,738 & 12,407 & 450 \\
    \midrule
    Weibo & 8,405 & 407,963 & 400 & 868 \\
    Reddit & 10,000 & 20,744,044 & 64 & 366 \\
    \bottomrule
    \end{tabular}%
  }
  \label{tab:stats}%
\end{table}%

Note that the anomaly generation in the synthetic datasets is following the anomaly injection method that has been widely used in previous research~\cite{liu2021anomaly,jin2021anemone,zheng2021generative,zhang2022subcr}, which is to generate a combined set of anomalies for each dataset by perturbing topological structure and nodal attributes, respectively. The detailed description of the anomaly injection is as follows.

\textbf{Injection of topological anomalies}: To obtain topological anomalies, the topological structure of networks is perturbed by generating small cliques composed of nodes that were originally not related.
    The insight is that in a small clique, a small group of nodes are significantly more interconnected with each other than the average, which can be considered a typical situation of topological anomalies in real-world graphs~\cite{liu2021anomaly}. 
    When generating a clique with the clique size $p$ and the number of cliques $q$, we randomly select $p$ nodes from the set of nodes $\gV$ and connect them fully. This implies that all the selected $p$ nodes are considered topological anomalies. To generate $q$ cliques, we repeat this process $q$ times. This results in a total of $p \times q$ topological anomalies. Following previous works, the value of $p$ is fixed as 15 and the value of $q$ is set to 5, 5, 15 for Cora, Citeseer, and Flickr, respectively.

\textbf{Injection of attributed anomalies}: We inject attributed anomalies by disturbing the attribute of nodes, which is introduced in~\cite{song2007conditional}. To generate an attributed anomaly, a node $v_{i}$ is randomly selected as the target, and then another $k$ nodes $(v_{1}^{c}, ..., v_{k}^{c})$ are sampled as a candidate set $\gV^{c}$. Next, we compute the Euclidean distance between the attribute vector $\mathbf{x}_{c}$ of each $v^{c}\in\gV^{c}$ and the attribute vector $\mathbf{x}_{i}$ of $v_{i}$. We then select the node $v_{j}^{c}\in \gV^{c}$ that has the largest Euclidean distance to $v_{i}$ and change $\mathbf{x}_{i}$ to $\mathbf{x}_{j}^{c}$. Following the previous works, the value of $k$ is set to 50.

\subsubsection{\textbf{Baselines}}
We compare the proposed \model with seventeen representative state-of-the-art unsupervised graph anomaly detection models, including five main groups: 

i) \textbf{Shallow Detection Models}: 
(1) SCAN~\cite{xu2007scan} is a clustering algorithm, which clusters vertices based on structural similarity to detect anomalies.
(2) MLPAE~\cite{sakurada2014anomaly} utilizes autoencoders onto both anomalous and benign data with a nonlinear MLP. 

ii) \textbf{Improved GNN-based Models}: 
(1) GAAN~\cite{chen2020generative} is a generative adversarial training framework with a GNN encoder to obtain real and fake node representations and a discriminator to recognize whether two connected nodes are from the real or fake graph.
(2) ALARM~\cite{peng2020deep} is a multi-view representation learning framework with multiple GNN encoders and a well-designed fusion operator between them.
(3) AAGNN~\cite{zhou2021subtractive} is an enhanced GNN, which utilizes subtractive aggregation to represent each node as the deviation from its neighbors. 

iii) \textbf{Graph AutoEncoder-based Models}: 
(1) GCNAE~\cite{kipf2016variational} is the GCN-based variational graph autoencoder and utilizes the reconstruction loss for anomaly detection.
(2) Dominant~\cite{ding2019deep} is a deep graph autoencoder-based method with a shared encoder. It detects the anomalies by computing the weighted sum of reconstruction error terms. 
(3) AnomalyDAE~\cite{fan2020anomalydae} is a dual graph autoencoder method with asymmetrical cross-modality interactions between structure and attribute.
(4) ComGA~\cite{luo2022comga} is a community-aware graph anomaly detection framework with a designed tailored deep GCN.

iv) \textbf{Graph Contrastive Learning Models}: 
(1) CoLA~\cite{liu2021anomaly} is a graph contrastive learning method that detects anomalies by evaluating the agreement between each node and its sampled subgraph.
(2) ANEMONE~\cite{jin2021anemone} is a multi-scale contrastive learning method, which captures the anomaly pattern by learning the agreements between nodes at both patch and context levels.
(3) SL-GAD~\cite{zheng2021generative} is self-supervised trained with generative and multi-view contrastive perspectives concurrently.
(4) Sub-CR~\cite{zhang2022subcr} employs the graph diffusion-based multi-view contrastive learning along with attribute reconstruction.

v) \textbf{Heterophily GNN-based Models}: 
(1) MixHop~\cite{abu2019mixhop} is a graph convolutional network with the mixed aggregation of multi-hop neighbors during one message passing.
(2) H2GCN~\cite{zhu2020beyond} is a heterophily graph model by the separate encoding of ego\&neighbor-embedding with higher-order neighbors.
(3) LINKX~\cite{lim2021large} is an MLP-based model with separate modeling for topology and features.
(4) GloGNN~\cite{li2022finding} is a method that considers both homophily and heterophily with the combination of low-pass and high-pass filters. We compare their autoencoder architectures for unsupervised learning.

\begin{table*}[thbp]
  \centering
  \caption{Unsupervised Graph anomaly detection comparison results on AUC and AP metrics (mean ± standard deviation in percentage over \textit{five} trial runs). The best and second-best results in each column are highlighted in \textbf{bold} font and \underline{underlined}.}
  \resizebox{\linewidth}{!}{
    \begin{tabular}{c|cc|cc|cc|cc|cc}
    \toprule
    \multirow{3}[6]{*}{\textbf{Model}} & \multicolumn{6}{c|}{\textbf{\textit{Synthetic Datasets}}}       & \multicolumn{4}{c}{\textbf{\textit{Real-world Datasets}}} \\
\cmidrule{2-11}          & \multicolumn{2}{c|}{\textbf{Cora}} & \multicolumn{2}{c|}{\textbf{Citeseer}} & \multicolumn{2}{c|}{\textbf{Flickr}} & \multicolumn{2}{c|}{\textbf{Weibo}} & \multicolumn{2}{c}{\textbf{Reddit}} \\
\cmidrule{2-11}          & \textbf{AUC}   & \textbf{AP}    & \textbf{AUC}   & \textbf{AP}    & \textbf{AUC}   & \textbf{AP}    & \textbf{AUC}   & \textbf{AP}    & \textbf{AUC}   & \textbf{AP} \\
    \midrule
    SCAN  & 0.6604\std{0.0163} & 0.0859\std{0.0044} & 0.6689\std{0.0136} & 0.0731\std{0.0032} & 0.6503\std{0.0120} & 0.3035\std{0.0227} & 0.7011\std{0.0000} &  0.1855\std{0.0000} & 0.4978\std{0.0000} & 0.0364\std{0.0000} \\
    MLPAE & 0.7565\std{0.0108} & 0.3528\std{0.0188} & 0.7396\std{0.0112} & 0.3124\std{0.0169} & 0.7466\std{0.0041} & 0.3484\std{0.0087} & 0.8946\std{0.0028} & 0.6696\std{0.0102} & 0.5108\std{0.0310} & 0.0359\std{0.0029} \\
    \midrule
    GAAN  & 0.7917\std{0.0118} & 0.3271\std{0.0124} & 0.8066\std{0.0036} & 0.3495\std{0.0101} & 0.7463\std{0.0041} & 0.3552\std{0.0119} & \underline{0.9249}\std{0.0000} & \underline{0.8104}\std{0.0000} & 0.5683\std{0.0001} & 0.0493\std{0.0001} \\
    ALARM & 0.8271\std{0.0223} & 0.2503\std{0.0379} & 0.8325\std{0.0121} & 0.3027\std{0.0559} & 0.6085\std{0.0034} & 0.0726\std{0.0013} & 0.9226\std{0.0000} & 0.8071\std{0.0000} & 0.5644\std{0.0003} & 0.0466\std{0.0001} \\
    AAGNN & 0.7590\std{0.0056} & 0.3744\std{0.0171} & 0.7202\std{0.0140} & 0.2345\std{0.0303} & 0.7454\std{0.0033} & 0.3506\std{0.0087} & 0.8066\std{0.0027} & 0.6679\std{0.0009} & 0.5442\std{0.0299} & 0.0400\std{0.0030} \\
    \midrule
    GCNAE & 0.7959\std{0.0104} & 0.3544\std{0.0350} & 0.7678\std{0.0114} & 0.3321\std{0.0243} & \underline{0.7471}\std{0.0056} & 0.3359\std{0.0125} & 0.8449\std{0.0032} & 0.5650\std{0.0030} & 0.5037\std{0.0015} & 0.0346\std{0.0001} \\
    Dominant & 0.8773\std{0.0134} & 0.3090\std{0.0438} & 0.8523\std{0.0051} & 0.3999\std{0.0092} & 0.6129\std{0.0035} & 0.0734\std{0.0013} & 0.8423\std{0.0117} & 0.6163\std{0.0237} & \underline{0.5752}\std{0.0056} & \underline{0.0570}\std{0.0019} \\
    AnomalyDAE & 0.8594\std{0.0068} & 0.3586\std{0.0148} & 0.8092\std{0.0059} & 0.3487\std{0.0103} & 0.7418\std{0.0051} & 0.3635\std{0.0107} & 0.8881\std{0.0165} & 0.6681\std{0.0874} & 0.4315\std{0.0001} & 0.0319\std{0.0000} \\
    ComGA & 0.7382\std{0.0162} & 0.1539\std{0.0242} & 0.7004\std{0.0150} & 0.0921\std{0.0085} & 0.6658\std{0.0033} & 0.1896\std{0.0120} & 0.9248\std{0.0006} & 0.8097\std{0.0009} & 0.4317\std{0.0001} & 0.0320\std{0.0001} \\
    \midrule
    MixHop & 0.7796\std{0.0107} & 0.2412\std{0.0300} & 0.7401\std{0.0122} & 0.3146\std{0.0195} & 0.7447\std{0.0060} & 0.3517\std{0.0229} & 0.8612\std{0.0018} & 0.6278\std{0.0061} & 0.5400\std{0.0175} & 0.0398\std{0.0028} \\
    H2GCN & 0.7827\std{0.0104} & 0.3479\std{0.0285} & 0.7361\std{0.0169} & 0.3064\std{0.0139} & 0.7463\std{0.0042} & 0.3526\std{0.0135} & 0.8546\std{0.0020} & 0.5604\std{0.0088} & 0.5476\std{0.0113} & 0.0401\std{0.0010} \\
    LINKX & 0.7601\std{0.0098} & 0.3605\std{0.0242} & 0.7416\std{0.0117} & 0.3187\std{0.0180} & 0.7466\std{0.0042} & \underline{0.3636}\std{0.0065} & 0.8018\std{0.0094} & 0.2822\std{0.0088} & 0.5576\std{0.0162} & 0.0421\std{0.0020} \\
    GloGNN & 0.7563\std{0.0083} & 0.3470\std{0.0269} & 0.7419\std{0.0124} & 0.3214\std{0.0220} & 0.7430\std{0.0013} & 0.3583\std{0.0054} & 0.9128\std{0.0251} & 0.7465\std{0.1280} & 0.5436\std{0.0364} & 0.0468\std{0.0085} \\
    \midrule
    CoLA  & 0.8887\std{0.0147} & 0.4804\std{0.0494} & 0.8211\std{0.0118} & 0.2302\std{0.0265} & 0.5612\std{0.0224} & 0.0694\std{0.0047} & 0.4842\std{0.0238} & 0.1006\std{0.0155} & 0.5149\std{0.0233} & 0.0393\std{0.0020} \\
    ANEMONE & 0.8966\std{0.0119} & \textbf{0.5425}\std{0.0489} & 0.8513\std{0.0159} & 0.3229\std{0.0212} & 0.5579\std{0.0271} & 0.0712\std{0.0062} & 0.3607\std{0.0120} & 0.0863\std{0.0147} & 0.4952\std{0.0188} & 0.0387\std{0.0026} \\
    SL-GAD & 0.8159\std{0.0238} & 0.3361\std{0.0337} & 0.7287\std{0.0201} & 0.2122\std{0.0273} & 0.7240\std{0.0087} & 0.3146\std{0.0123} & 0.4298\std{0.0073} & 0.0899\std{0.0036} & 0.5488\std{0.0142} & 0.0424\std{0.0032} \\
    Sub-CR & \underline{0.8968}\std{0.0118} & 0.4771\std{0.0102} & \underline{0.9060}\std{0.0095} & \underline{0.4751}\std{0.0254} & 0.7423\std{0.0038} & 0.3611\std{0.0145} & 0.6404\std{0.0070} & 0.4900\std{0.0103} & 0.5327\std{0.0156} & 0.0379\std{0.0013} \\
    \midrule
    \textbf{\model} (ours) & \textbf{0.9689}\std{0.0013} & \underline{0.4981}\std{0.0196} & \textbf{0.9695}\std{0.0047} & \textbf{0.5451}\std{0.0110} & \textbf{0.7625}\std{0.0054} & \textbf{0.3673}\std{0.0096} & \textbf{0.9805}\std{0.0011} & \textbf{0.8639}\std{0.0128} & \textbf{0.6022}\std{0.0079} & \textbf{0.0629}\std{0.0027} \\
    \textit{Improv.} (\%) & +8.04\% & --- & +7.01\% & +14.73\% & +2.06\% & +1.02\% & +6.01\% & +6.60\% & +4.69\% & +10.35\% \\
    \bottomrule
    \end{tabular}%
    }
  \label{tab:main_tab}%
\end{table*}%

\subsubsection{\textbf{Evaluation Metrics}} 
We evaluate the models with AUC and AP, the widely-adopted metrics in previous anomaly detection works~\cite{ding2019deep,tang2022rethinking}, to evaluate the detection performance. The higher AUC and AP values indicate better anomaly detection performance. Note that we run all the experiments \textit{five} times with different random seeds and report the average results with standard deviation to prevent extreme cases.

\subsubsection{\textbf{Hyper-parameter Settings}}
The embedding size is fixed to 64 and the embedding parameters are initialized with the Xavier method. The loss function is optimized with Adam optimizer. The learning rate of \model is searched from \{$5\times 10^{-2}, 1 \times 10^{-2}, 5 \times 10^{-3}, 1 \times 10^{-3}$\}. For all baselines, we retain the parameter settings in their corresponding papers to keep the comparison fair. All experiments are conducted on the Centos system equipped with NVIDIA RTX-3090 GPUs.

\subsection{Main Results}
In this subsection, we compare \model with seventeen state-of-the-art baselines. The comparison results are reported in Table \ref{tab:main_tab}. From these results, we have the following observation:
\begin{itemize}[leftmargin=*]
    \item \textbf{\model can achieve significant improvement over state-of-the-art models on both synthetic and real-world datasets}. From the table, we observe that \model achieves the best AUC metric in all datasets and the optimal AP metric in 4 out of the 5 datasets. Specifically, for the AUC metric, \model outperforms the best baseline by 8.04\%, 7.01\%, 2.06\%, 6.01\%, and 4.69\% on Cora, Citeseer, Flickr, Weibo, and Reddit, respectively, which brings the gains of 5.56\% on average. 
    These results verify the effectiveness of \model in distinguishing representation learning and better anomaly detection performance.
    \item \textbf{The performance difference of baseline models on real-world datasets is more pronounced compared to that on synthetic datasets}, which exhibits the same observation with~\cite{liu2022bond} and shows that detecting real-world anomalies is more challenging. This may be because anomalies in the real world are more diverse, while the patterns of injected anomalies in the synthetic datasets are relatively uniform due to the fixed injection method.
    Therefore, in the face of diverse real-world anomaly patterns, the design of adaptive solutions is more appropriate.
    \item \textbf{Graph contrastive learning methods perform well on synthetic datasets but relatively poorly on real-world datasets}. This suggests that using sole contrastive schemes with simple vanilla GNN encoders cannot fully explore the diverse abnormal patterns in the real world. The heterophilic connections brought by real-world anomalies are more diverse, and encoding nodes indiscriminately through the vanilla GNN would further weaken the sample quality of contrastive learning.
\end{itemize}

\begin{table}[tbp]
  \centering
  \caption{Ablation study results compared with two \model variants.}
  \resizebox{0.96\linewidth}{!}{
    \begin{tabular}{c|ccc}
    \toprule
    Variant & Flick & Weibo & Reddit \\
    \midrule
    \textbf{\model} & \textbf{0.7625}\std{0.0054} & \textbf{0.9805}\std{0.0011} & \textbf{0.6022}\std{0.0079} \\
    \midrule
    \model-\textit{w/o TR} &  0.7475\std{0.0041}   &  0.8698\std{0.0093}     &  0.5801\std{0.0096} \\
    \model-\textit{w/o AR} &  0.5937\std{0.0252}   &   0.9725\std{0.0015}    &  0.6019\std{0.0052} \\
    \bottomrule
    \end{tabular}%
    }
  \label{tab:ablation}%
\end{table}%

\subsection{Ablation Study}
To verify the effectiveness of the design of \model, we conduct various ablation studies on \model.
First, we conduct the component ablation study on \model with its two component variants: 
(1) \model-\textit{w/o TR} removes the topology reconstruction module for both training and anomaly scoring.
(2) \model-\textit{w/o AR} removes the attribute reconstruction for both training and anomaly scoring.
The study results can be found in Table \ref{tab:ablation}, from which we have the following observations:
First, compared to \textit{w/o TR} and \textit{w/o AR} variants, \model gains significant improvements which proves the necessity and effectiveness of these modules.
Second, the impact of \textit{w/o TR} and \textit{w/o AR} varies from the datasets, which is related to the diversity of anomalies on different datasets.

\begin{figure}[tbp]
    \centering
    \includegraphics[width=\linewidth, trim=0cm 0cm 0cm 0cm,clip]{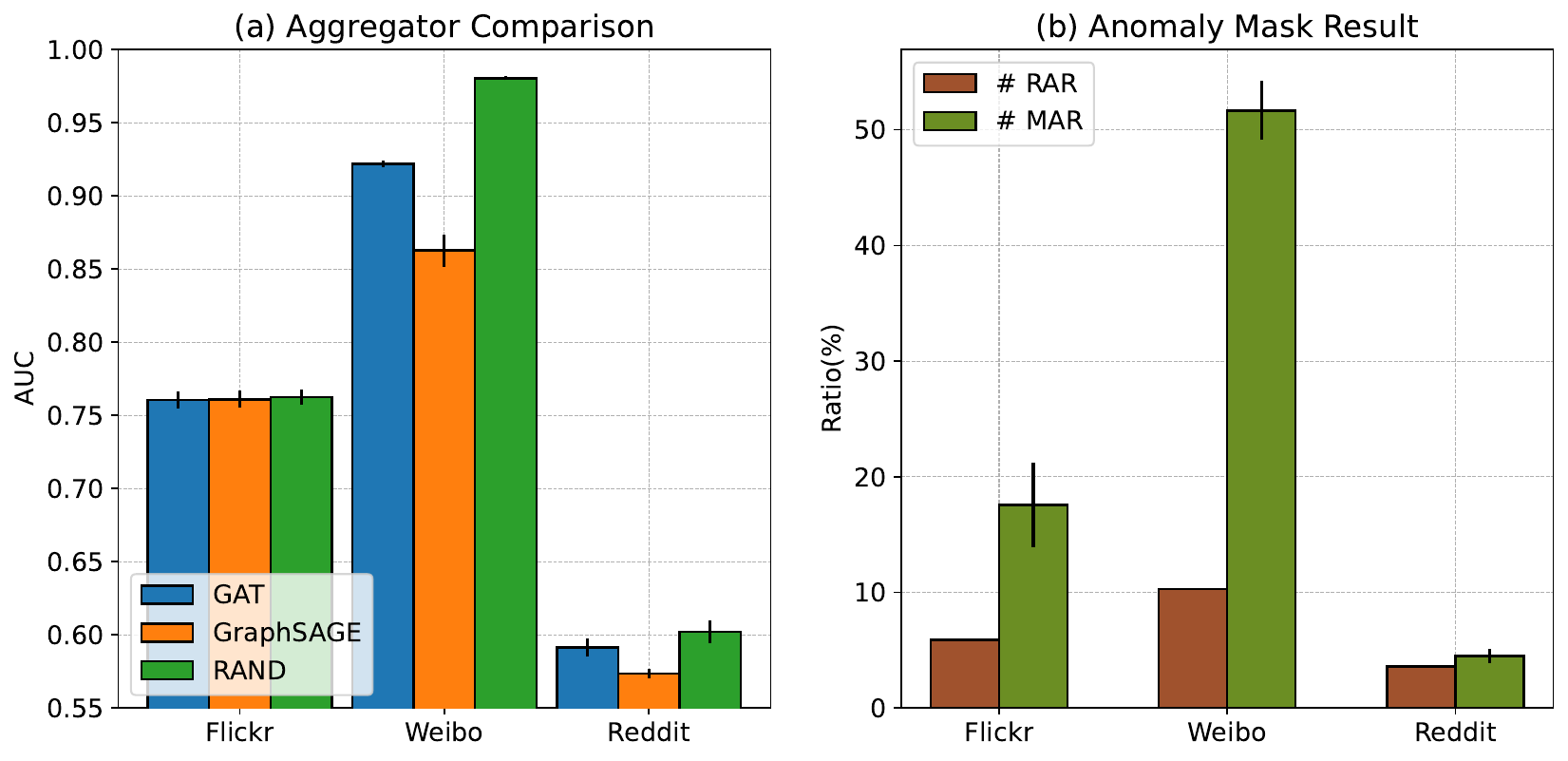}
    \caption{(a): Ablation study to analyze the effect of designed message aggregator of \model. (b): Study of the anomaly mask result in Eq.(\ref{eq:mask_ids}), where \textbf{\#RAR} denotes the \textbf{raw anomaly ratio} of each dataset and \textbf{\#MAR} denotes the \textbf{masked anomaly ratio} by \model with the mask rate of 3\%.}
    \label{fig:agg_mask}
\end{figure}

\begin{figure}[tbp]
    \centering
    \includegraphics[width=\linewidth, trim=0cm 0cm 0cm 0cm,clip]{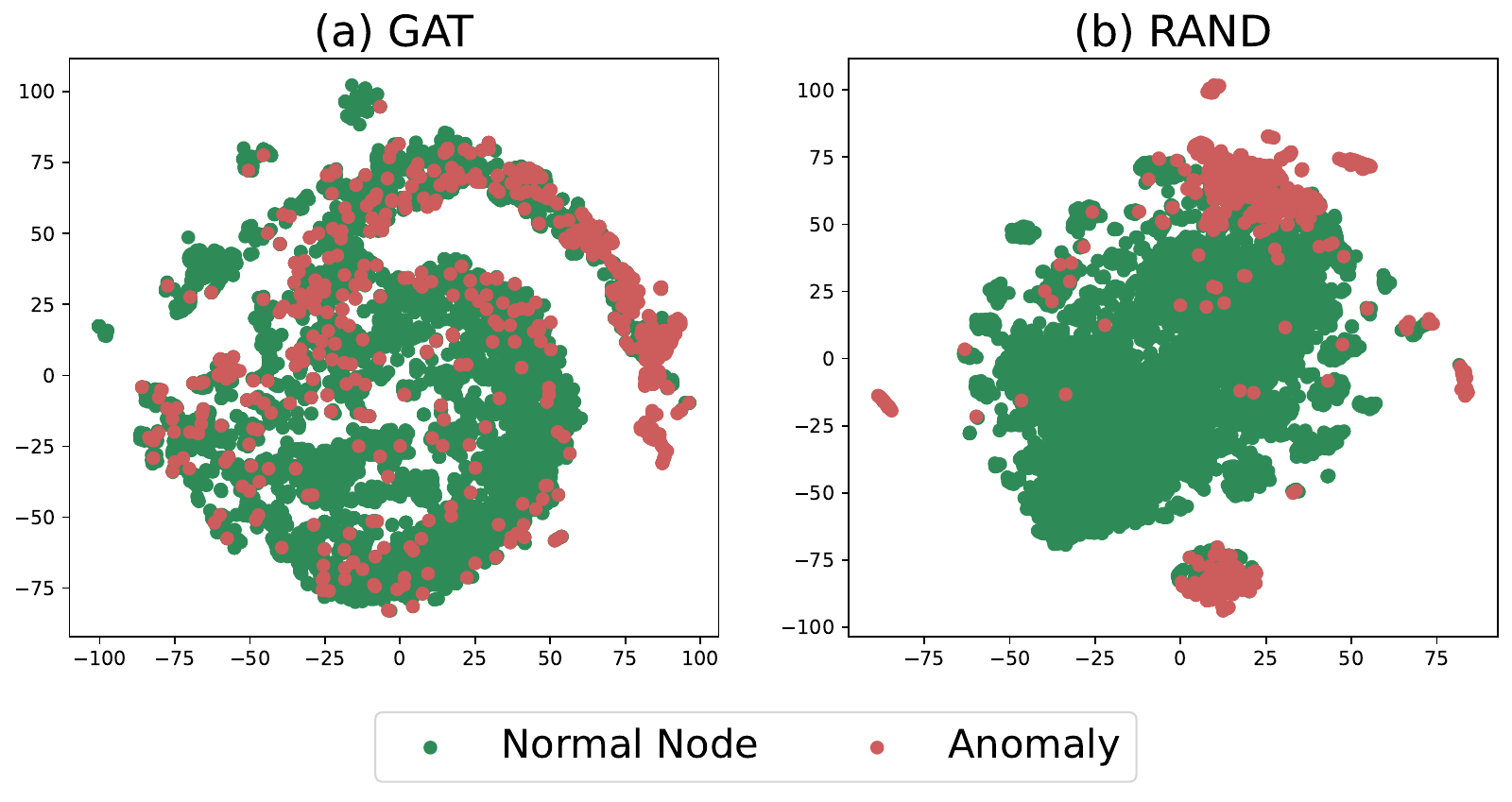}
    \caption{The visualization of node representations from the trained models. The message passing of our proposed \model can learn more distinguishable embeddings between normal nodes and anomalies in the latent space.}
    \label{fig:emb_comp}
    \vspace{-0.8em}
\end{figure}

Furthermore, we conduct the ablation study on the message aggregator of \model by comparing it with GraphSAGE~\cite{hamilton2017inductive} and GAT~\cite{veličković2018graph}. Figure \ref{fig:agg_mask}-(a) illustrates the study results. From the results, we can find that the designed message aggregator achieves the best detection performance. Furthermore, GAT performs relatively better results than GraphSAGE, the possible reason is that the attention mechanism in GAT helps it discriminate the anomalies. In Figure \ref{fig:emb_comp} we further conduct the node representation visualization through the widely-used T-SNE~\cite{van2008visualizing} method, and from it we can find that the aggregator of \model has a significantly greater discriminatory ability for anomalies.
Then, in Figure \ref{fig:agg_mask}-(b), we study the anomaly mask results of \model under the set mask rate of 3\%. We can observe that by using only a 3\% masking rate, it is possible to mask the ratio of anomalies that are much higher than the raw anomaly rate in the dataset (in Weibo, the masked anomaly rate even exceeds 50\%), which helps \model to widen the gap between representations of normal nodes and anomalies.

\subsection{Case Study}
\begin{figure}[tbp]
    \centering
    \includegraphics[width=\linewidth, trim=0cm 0cm 0cm 0cm,clip]{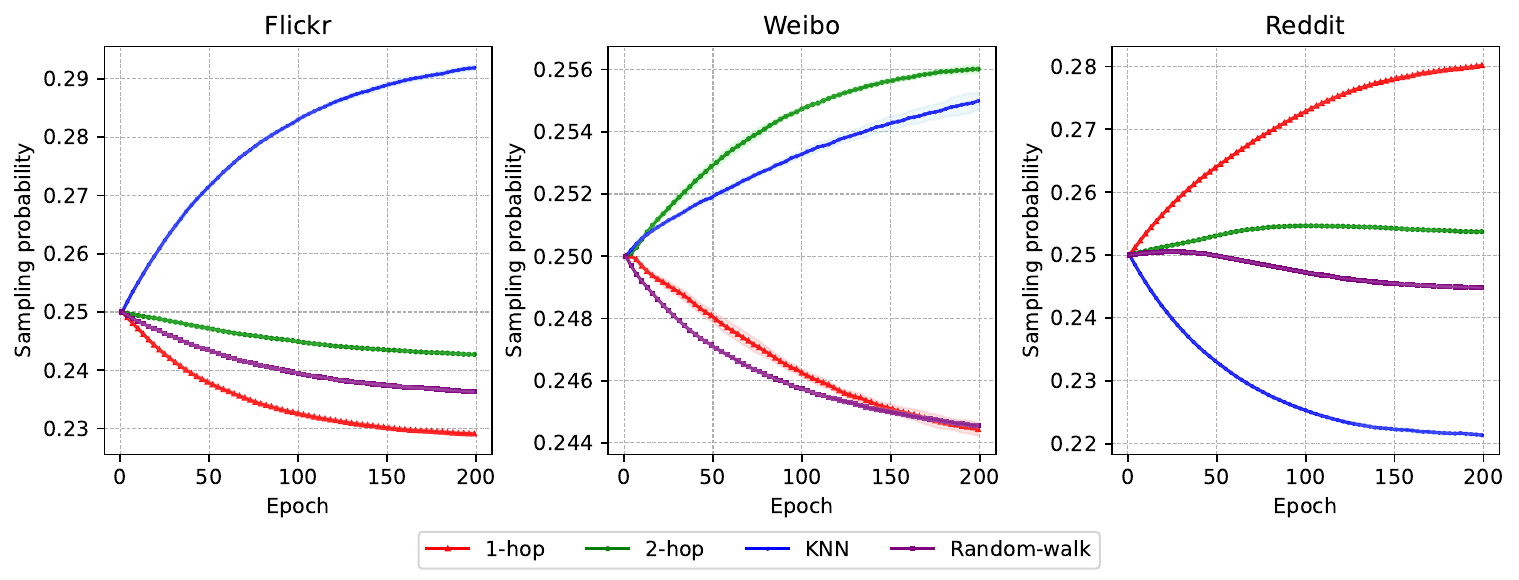}
    \caption{Case study of the dynamic selection probability changing during the training iterations.}
    \label{fig:case_study}
\end{figure}

We further conduct the case study of the dynamic changing of the selection probabilities of different strategies during the training process, of which the results are shown in Figure \ref{fig:case_study}.

From the table, we can find that \model has different preferences for different types of neighbors on different datasets, which is due to the diverse characteristics of the graphs and their anomalies. For example, on Flickr, \model prefers to aggregate KNN neighbors, which indicates the structurally defined neighbors are not so effective for it. This conclusion is consistent with the result that \textit{w/o AR} has a greater impact than \textit{w/o TR} in the ablation study on Flickr, indicating that the adaptive neighbor selection we designed is helpful.

\subsection{Parameter Study}
\subsubsection{\textbf{Effect of the balanced parameter}}
\begin{figure}[tbp]
    \centering
    \includegraphics[width=\linewidth, trim=0cm 0cm 0cm 0cm,clip]{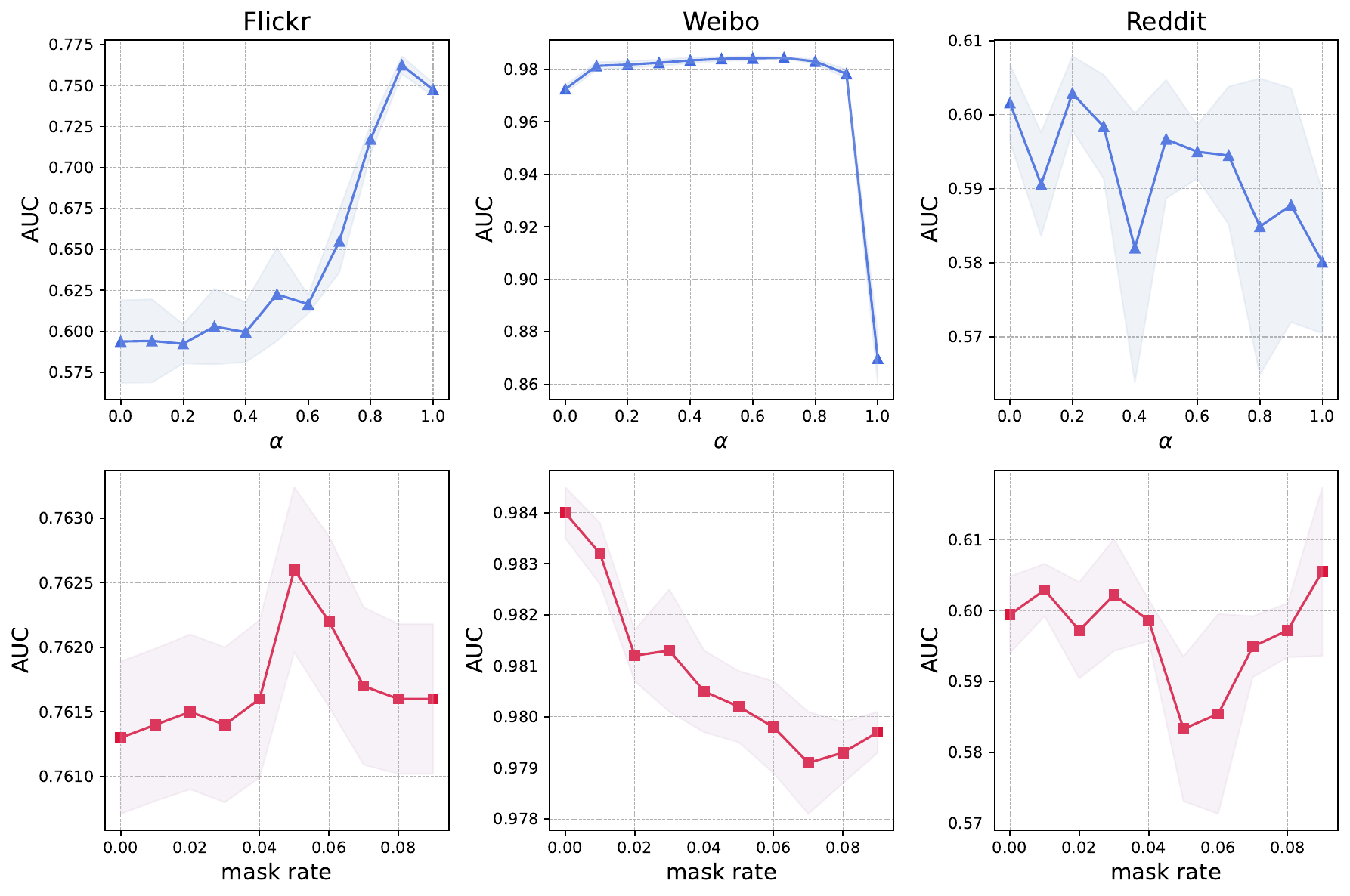}
    \caption{Parameter study results on the balanced hyper-parameter $\alpha$ and mask rate of Flick, Weibo, and Reddit datasets on the AUC metric.}
    \label{fig:param_analysis}
    \vspace{-0.8em}
\end{figure}

We investigate the effectiveness of the changing of the balanced parameter $\alpha$ on \model from 0.0 to 1.0 with a step of 0.1. 
The study results are shown in the first row of Figure \ref{fig:param_analysis}.
From the results, we observe that a suitable value of $\alpha$ is important for the model performance, which varies from different datasets, e.g. a related large value of $\alpha$ is better for Flickr while a small value is better for Reddit.

\subsubsection{\textbf{Effect of the mask rate}}
We also study the effect of the mask rate by varying it from the range of [0, 0.09].
The study results are shown in the second row of Figure \ref{fig:param_analysis}.
From the results, we can find that the suitable mask rate varies from different datasets.
Generally, adding a few node masking can bring performance gains to the model, but it is without positive gains on Weibo though the masked anomaly ratio is high.

%% file: conclusion.tex
In this paper, we propose \model, a novel method for unsupervised graph anomaly detection with reinforcement neighborhood selection.
\model first extends the candidate neighborhoods and then adaptively selects proper neighbors by reinforcement learning.
Furthermore, \model introduces a more anomaly-distinguishing message aggregator for passing more consistent messages for central nodes in an unsupervised manner.
Extensive experiments on both synthetic and real-world datasets illustrate that our proposed \model achieves state-of-the-art unsupervised graph anomaly detection performance.
In future works, we will explore more efficient neighborhood selection schemes to make the model more suitable for anomaly detection on large-scale graphs.

%% file: main_cr.bbl
\begin{thebibliography}{10}
\providecommand{\url}[1]{#1}
\csname url@samestyle\endcsname
\providecommand{\newblock}{\relax}
\providecommand{\bibinfo}[2]{#2}
\providecommand{\BIBentrySTDinterwordspacing}{\spaceskip=0pt\relax}
\providecommand{\BIBentryALTinterwordstretchfactor}{4}
\providecommand{\BIBentryALTinterwordspacing}{\spaceskip=\fontdimen2\font plus
\BIBentryALTinterwordstretchfactor\fontdimen3\font minus
  \fontdimen4\font\relax}
\providecommand{\BIBforeignlanguage}[2]{{%
\expandafter\ifx\csname l@#1\endcsname\relax
\typeout{** WARNING: IEEEtran.bst: No hyphenation pattern has been}%
\typeout{** loaded for the language `#1'. Using the pattern for}%
\typeout{** the default language instead.}%
\else
\language=\csname l@#1\endcsname
\fi
#2}}
\providecommand{\BIBdecl}{\relax}
\BIBdecl

\bibitem{ding2019deep}
K.~Ding, J.~Li, R.~Bhanushali, and H.~Liu, ``Deep anomaly detection on
  attributed networks,'' in \emph{Proceedings of the 2019 SIAM International
  Conference on Data Mining}.\hskip 1em plus 0.5em minus 0.4em\relax SIAM,
  2019, pp. 594--602.

\bibitem{dagad}
F.~Liu, X.~Ma, J.~Wu, J.~Yang, S.~Xue, A.~Beheshti, C.~Zhou, H.~Peng, Q.~Z.
  Sheng, and C.~C. Aggarwal, ``Dagad: Data augmentation for graph anomaly
  detection,'' in \emph{2022 IEEE International Conference on Data Mining
  (ICDM)}, 2022, pp. 259--268.

\bibitem{wang2019semi}
D.~Wang, J.~Lin, P.~Cui, Q.~Jia, Z.~Wang, Y.~Fang, Q.~Yu, J.~Zhou, S.~Yang, and
  Y.~Qi, ``A semi-supervised graph attentive network for financial fraud
  detection,'' in \emph{2019 IEEE International Conference on Data Mining
  (ICDM)}.\hskip 1em plus 0.5em minus 0.4em\relax IEEE, 2019, pp. 598--607.

\bibitem{sun2022rumor}
T.~Sun, Z.~Qian, S.~Dong, P.~Li, and Q.~Zhu, ``Rumor detection on social media
  with graph adversarial contrastive learning,'' in \emph{Proceedings of the
  ACM Web Conference}, 2022, pp. 2789--2797.

\bibitem{zhou2021hierarchical}
X.~Zhou, W.~Liang, W.~Li, K.~Yan, S.~Shimizu, I.~Kevin, and K.~Wang,
  ``Hierarchical adversarial attacks against graph-neural-network-based iot
  network intrusion detection system,'' \emph{IEEE Internet of Things Journal},
  vol.~9, no.~12, pp. 9310--9319, 2021.

\bibitem{kipf2016semi}
T.~N. Kipf and M.~Welling, ``Semi-supervised classification with graph
  convolutional networks,'' in \emph{ICLR}, 2017.

\bibitem{hamilton2017inductive}
W.~Hamilton, Z.~Ying, and J.~Leskovec, ``Inductive representation learning on
  large graphs,'' \emph{Advances in neural information processing systems},
  vol.~30, 2017.

\bibitem{ma2021comprehensive}
X.~Ma, J.~Wu, S.~Xue, J.~Yang, C.~Zhou, Q.~Z. Sheng, H.~Xiong, and L.~Akoglu,
  ``A comprehensive survey on graph anomaly detection with deep learning,''
  \emph{IEEE Transactions on Knowledge and Data Engineering}, 2021.

\bibitem{mcpherson2001birds}
M.~McPherson, L.~Smith-Lovin, and J.~M. Cook, ``Birds of a feather: Homophily
  in social networks,'' \emph{Annual review of sociology}, pp. 415--444, 2001.

\bibitem{dou2020enhancing}
Y.~Dou, Z.~Liu, L.~Sun, Y.~Deng, H.~Peng, and P.~S. Yu, ``Enhancing graph
  neural network-based fraud detectors against camouflaged fraudsters,'' in
  \emph{CIKM}, 2020, pp. 315--324.

\bibitem{liu2022bond}
K.~Liu, Y.~Dou, Y.~Zhao, X.~Ding, X.~Hu, R.~Zhang, K.~Ding, C.~Chen, H.~Peng,
  K.~Shu \emph{et~al.}, ``Bond: Benchmarking unsupervised outlier node
  detection on static attributed graphs,'' in \emph{NIPS Datasets and
  Benchmarks Track}, 2022.

\bibitem{xu2007scan}
X.~Xu, N.~Yuruk, Z.~Feng, and T.~A. Schweiger, ``Scan: a structural clustering
  algorithm for networks,'' in \emph{KDD}, 2007, pp. 824--833.

\bibitem{sakurada2014anomaly}
M.~Sakurada and T.~Yairi, ``Anomaly detection using autoencoders with nonlinear
  dimensionality reduction,'' in \emph{MLSDA 2nd workshop on machine learning
  for sensory data analysis}, 2014, pp. 4--11.

\bibitem{chen2020generative}
Z.~Chen, B.~Liu, M.~Wang, P.~Dai, J.~Lv, and L.~Bo, ``Generative adversarial
  attributed network anomaly detection,'' in \emph{CIKM}, 2020, pp. 1989--1992.

\bibitem{peng2020deep}
Z.~Peng, M.~Luo, J.~Li, L.~Xue, and Q.~Zheng, ``A deep multi-view framework for
  anomaly detection on attributed networks,'' \emph{IEEE Transactions on
  Knowledge and Data Engineering}, 2020.

\bibitem{zhou2021subtractive}
S.~Zhou, Q.~Tan, Z.~Xu, X.~Huang, and F.-l. Chung, ``Subtractive aggregation
  for attributed network anomaly detection,'' in \emph{CIKM}, 2021, pp.
  3672--3676.

\bibitem{kipf2016variational}
T.~N. Kipf and M.~Welling, ``Variational graph auto-encoders,'' \emph{arXiv
  preprint arXiv:1611.07308}, 2016.

\bibitem{fan2020anomalydae}
H.~Fan, F.~Zhang, and Z.~Li, ``Anomalydae: Dual autoencoder for anomaly
  detection on attributed networks,'' in \emph{ICASSP}.\hskip 1em plus 0.5em
  minus 0.4em\relax IEEE, 2020, pp. 5685--5689.

\bibitem{luo2022comga}
X.~Luo, J.~Wu, A.~Beheshti, J.~Yang, X.~Zhang, Y.~Wang, and S.~Xue, ``Comga:
  Community-aware attributed graph anomaly detection,'' in \emph{WSDM}, 2022,
  pp. 657--665.

\bibitem{liu2021anomaly}
Y.~Liu, Z.~Li, S.~Pan, C.~Gong, C.~Zhou, and G.~Karypis, ``Anomaly detection on
  attributed networks via contrastive self-supervised learning,'' \emph{IEEE
  transactions on neural networks and learning systems}, 2021.

\bibitem{jin2021anemone}
M.~Jin, Y.~Liu, Y.~Zheng, L.~Chi, Y.-F. Li, and S.~Pan, ``Anemone: Graph
  anomaly detection with multi-scale contrastive learning,'' in \emph{CIKM},
  2021, pp. 3122--3126.

\bibitem{zheng2021generative}
Y.~Zheng, M.~Jin, Y.~Liu, L.~Chi, K.~T. Phan, and Y.-P.~P. Chen, ``Generative
  and contrastive self-supervised learning for graph anomaly detection,''
  \emph{IEEE Transactions on Knowledge and Data Engineering}, 2021.

\bibitem{zhang2022subcr}
J.~Zhang, S.~Wang, and S.~Chen, ``Reconstruction enhanced multi-view
  contrastive learning for anomaly detection on attributed networks.'' in
  \emph{IJCAI}, 2022, pp. 2376--2382.

\bibitem{veličković2018graph}
P.~Veličković, G.~Cucurull, A.~Casanova, A.~Romero, P.~Liò, and Y.~Bengio,
  ``Graph attention networks,'' in \emph{ICLR}, 2018.

\bibitem{abu2019mixhop}
S.~Abu-El-Haija, B.~Perozzi, A.~Kapoor, N.~Alipourfard, K.~Lerman,
  H.~Harutyunyan, G.~Ver~Steeg, and A.~Galstyan, ``Mixhop: Higher-order graph
  convolutional architectures via sparsified neighborhood mixing,'' in
  \emph{ICML}.\hskip 1em plus 0.5em minus 0.4em\relax PMLR, 2019, pp. 21--29.

\bibitem{zhu2020beyond}
J.~Zhu, Y.~Yan, L.~Zhao, M.~Heimann, L.~Akoglu, and D.~Koutra, ``Beyond
  homophily in graph neural networks: Current limitations and effective
  designs,'' \emph{Advances in Neural Information Processing Systems}, vol.~33,
  pp. 7793--7804, 2020.

\bibitem{lim2021large}
D.~Lim, F.~Hohne, X.~Li, S.~L. Huang, V.~Gupta, O.~Bhalerao, and S.~N. Lim,
  ``Large scale learning on non-homophilous graphs: New benchmarks and strong
  simple methods,'' \emph{Advances in Neural Information Processing Systems},
  vol.~34, pp. 20\,887--20\,902, 2021.

\bibitem{li2022finding}
X.~Li, R.~Zhu, Y.~Cheng, C.~Shan, S.~Luo, D.~Li, and W.~Qian, ``Finding global
  homophily in graph neural networks when meeting heterophily,'' in
  \emph{ICML}.\hskip 1em plus 0.5em minus 0.4em\relax PMLR, 2022, pp.
  13\,242--13\,256.

\bibitem{shi2022h2}
F.~Shi, Y.~Cao, Y.~Shang, Y.~Zhou, C.~Zhou, and J.~Wu, ``H2-fdetector: a
  gnn-based fraud detector with homophilic and heterophilic connections,'' in
  \emph{Proceedings of the ACM Web Conference 2022}, 2022, pp. 1486--1494.

\bibitem{kober2013reinforcement}
J.~Kober, J.~A. Bagnell, and J.~Peters, ``Reinforcement learning in robotics: A
  survey,'' \emph{The International Journal of Robotics Research}, vol.~32,
  no.~11, pp. 1238--1274, 2013.

\bibitem{lanctot2017unified}
M.~Lanctot, V.~Zambaldi, A.~Gruslys, A.~Lazaridou, K.~Tuyls, J.~P{\'e}rolat,
  D.~Silver, and T.~Graepel, ``A unified game-theoretic approach to multiagent
  reinforcement learning,'' \emph{Advances in neural information processing
  systems}, vol.~30, 2017.

\bibitem{nie2023reinforcement}
M.~Nie, D.~Chen, and D.~Wang, ``Reinforcement learning on graphs: A survey,''
  \emph{IEEE Transactions on Emerging Topics in Computational Intelligence},
  2023.

\bibitem{lai2020policy}
K.-H. Lai, D.~Zha, K.~Zhou, and X.~Hu, ``Policy-gnn: Aggregation optimization
  for graph neural networks,'' in \emph{KDD}, 2020, pp. 461--471.

\bibitem{ansgt}
Z.~ZHANG, Q.~Liu, Q.~Hu, and C.-K. Lee, ``Hierarchical graph transformer with
  adaptive node sampling,'' in \emph{Advances in Neural Information Processing
  Systems}, S.~Koyejo, S.~Mohamed, A.~Agarwal, D.~Belgrave, K.~Cho, and A.~Oh,
  Eds., vol.~35, 2022, pp. 21\,171--21\,183.

\bibitem{dai2022towards}
E.~Dai, W.~Jin, H.~Liu, and S.~Wang, ``Towards robust graph neural networks for
  noisy graphs with sparse labels,'' in \emph{WSDM}, 2022, pp. 181--191.

\bibitem{auer2002nonstochastic}
P.~Auer, N.~Cesa-Bianchi, Y.~Freund, and R.~E. Schapire, ``The nonstochastic
  multiarmed bandit problem,'' \emph{SIAM journal on computing}, vol.~32,
  no.~1, pp. 48--77, 2002.

\bibitem{wang2020personalized}
H.~Wang, Z.~Wei, J.~Gan, S.~Wang, and Z.~Huang, ``Personalized pagerank to a
  target node, revisited,'' in \emph{KDD}, 2020, pp. 657--667.

\bibitem{akoglu2015graph}
L.~Akoglu, H.~Tong, and D.~Koutra, ``Graph based anomaly detection and
  description: a survey,'' \emph{Data mining and knowledge discovery}, vol.~29,
  no.~3, pp. 626--688, 2015.

\bibitem{sen2008collective}
P.~Sen, G.~Namata, M.~Bilgic, L.~Getoor, B.~Galligher, and T.~Eliassi-Rad,
  ``Collective classification in network data,'' \emph{AI magazine}, vol.~29,
  no.~3, pp. 93--93, 2008.

\bibitem{tang2009relational}
L.~Tang and H.~Liu, ``Relational learning via latent social dimensions,'' in
  \emph{KDD}, 2009, pp. 817--826.

\bibitem{zhao2020error}
T.~Zhao, C.~Deng, K.~Yu, T.~Jiang, D.~Wang, and M.~Jiang, ``Error-bounded graph
  anomaly loss for gnns,'' in \emph{CIKM}, 2020, pp. 1873--1882.

\bibitem{kumar2019predicting}
S.~Kumar, X.~Zhang, and J.~Leskovec, ``Predicting dynamic embedding trajectory
  in temporal interaction networks,'' in \emph{KDD}, 2019, pp. 1269--1278.

\bibitem{song2007conditional}
X.~Song, M.~Wu, C.~Jermaine, and S.~Ranka, ``Conditional anomaly detection,''
  \emph{IEEE Transactions on knowledge and Data Engineering}, vol.~19, no.~5,
  pp. 631--645, 2007.

\bibitem{tang2022rethinking}
J.~Tang, J.~Li, Z.~Gao, and J.~Li, ``Rethinking graph neural networks for
  anomaly detection,'' in \emph{ICML}, 2022.

\bibitem{van2008visualizing}
L.~Van~der Maaten and G.~Hinton, ``Visualizing data using t-sne.''
  \emph{Journal of machine learning research}, vol.~9, no.~11, 2008.

\end{thebibliography}
